\documentclass[final]{cvpr}
\usepackage{times}
\usepackage{epsfig}
\usepackage{graphicx}
\usepackage{amsmath}
\usepackage{amssymb}

\usepackage{array,booktabs,calc}
\usepackage{placeins}
\usepackage[inline, shortlabels]{enumitem}
\usepackage{graphicx}
\usepackage{multirow}
\usepackage{wrapfig}
\usepackage{paralist}
\usepackage{color}
\usepackage{colortbl}
\usepackage[dvipsnames]{xcolor}
\usepackage{inconsolata}
\usepackage{floatrow}
\usepackage{gensymb}
\usepackage[backref]{hyperref}
\usepackage{makecell}

\setlist{itemjoin ={,\enspace},itemjoin* = { and\enspace}}
\definecolor{citecolor}{HTML}{0071bc}
\hypersetup{
  breaklinks   = true,
  colorlinks   = true, %
  urlcolor     = RubineRed, %
  linkcolor    = RubineRed, %
  citecolor    = citecolor %
}
\newcommand{\PAR}[1]{\vskip4pt \noindent{\bf #1~}}

\def\cvprPaperID{3433} %
\def\confYear{CVPR 2021}
\newcommand{\shortname}{NeuralRecon\xspace}

\newcommand\blfootnote[1]{%
  \begingroup
  \renewcommand\thefootnote{}\footnote{#1}%
  \addtocounter{footnote}{-1}%
  \endgroup
}
\hypersetup{pdfnewwindow}
\newcommand{\urlNewWindow}[1]{\href[pdfnewwindow=true]{#1}{\nolinkurl{#1}}}

\begin{document}

\title{\shortname: Real-Time Coherent 3D Reconstruction from Monocular Video}

\author{
    Jiaming Sun$^{1,2*}$ 
    \quad Yiming Xie$^{1*}$ 
    \quad Linghao Chen$^{1}$ 
    \quad Xiaowei Zhou$^{1}$
    \quad Hujun Bao$^{1\dagger}$ 
    \\
    $^1$Zhejiang University \quad 
    $^2$SenseTime Research \quad
}

\maketitle

\begin{abstract}
  We present a novel framework named \shortname for real-time 3D scene reconstruction from a monocular video.
Unlike previous methods that estimate single-view depth maps separately on each key-frame and fuse them later, we propose to directly reconstruct local surfaces represented as sparse TSDF volumes for each video fragment sequentially by a neural network. A learning-based TSDF fusion module based on gated recurrent units is used to guide the network to fuse features from previous fragments. 
This design allows the network to capture local smoothness prior and global shape prior of 3D surfaces when sequentially reconstructing the surfaces, resulting in accurate, coherent, and real-time surface reconstruction.     
The experiments on ScanNet and 7-Scenes datasets show that our system outperforms state-of-the-art methods in terms of both accuracy and speed.
To the best of our knowledge, this is the first learning-based system that is able to reconstruct dense coherent 3D geometry in real-time. 
Code is available at the project page: \urlNewWindow{https://zju3dv.github.io/neuralrecon/}.

\end{abstract}

\blfootnote{$^*$The first two authors contributed equally. The authors are affiliated with the State Key Lab of CAD\&CG and ZJU-SenseTime Joint Lab of 3D Vision. $^\dagger$Corresponding author: Hujun Bao.}
\section{Introduction}\label{sec:intro}
3D scene reconstruction is one of the central tasks in 3D computer vision with many applications. 
In augmented reality (AR) for example, to enable realistic and immersive interactions between AR effects and the surrounding physical scene, 
3D reconstruction needs to be accurate, coherent and performed in real-time.
While camera motion can be tracked accurately with state-of-the-art visual-inertial SLAM systems \cite{camposORBSLAM3AccurateOpenSource2020,qinGeneralOptimizationbasedFramework2019a, AugmentedRealityApple}, 
real-time image-based dense reconstruction remains to be a challenging problem due to low reconstruction quality and high computation demands.
\begin{figure}[tb]
    \centering
    \includegraphics[width=1\linewidth]{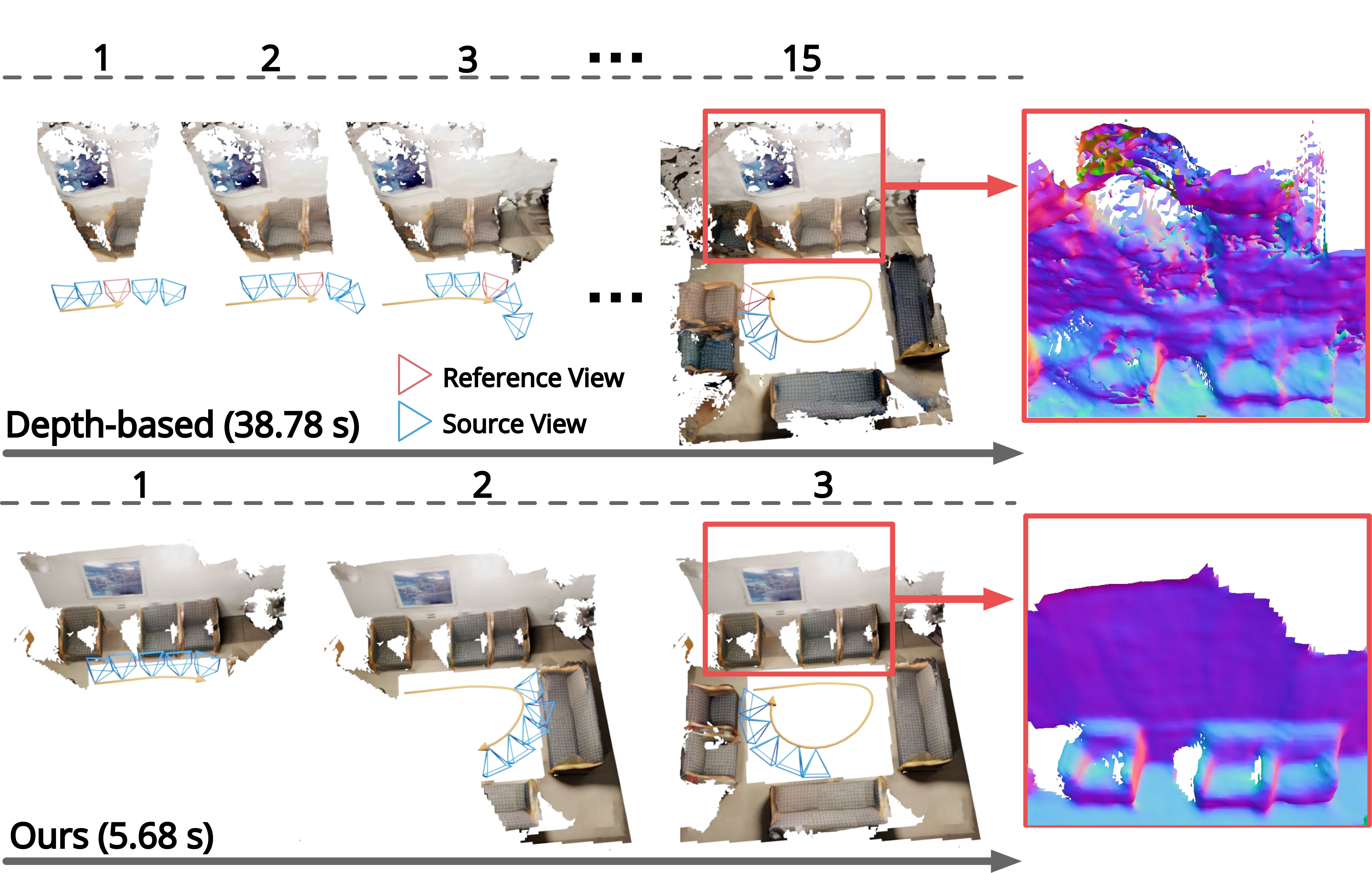}
    \caption{
        \textbf{Comparison between depth-based 3D reconstruction methods and the proposed method.} 
        In depth-based methods, key-frame depths are estimated separately from each key frame, and later fused into a TSDF volume.
        In the proposed method, the TSDF volume is directly predicted with all the key frames in a local window.
        This design leads to a much more coherent reconstruction and real-time speed. 
    }
    \label{fig:teaser}
\end{figure}

Most image-based real-time 3D reconstruction pipelines \cite{schops3DModelingGo2015,yangRealtimeMonocularDense2017} adopt the depth map fusion approach, which resemble RGB-D reconstruction methods like KinectFusion \cite{newcombeKinectFusionRealTimeDense2011}.
Single-view depth maps from each key frame are first estimated with real-time multi-view depth estimation methods like \cite{wangMVDepthNetRealtimeMultiview2018,liuNeuralRGBSensing2019,houMultiViewStereoTemporal2019,valentinDepthMotionSmartphone2019}. 
The estimated depth maps are later filtered with criteria like multi-view consistency and temporal smoothness, and fused into a Truncated Signed Distance Function (TSDF) volume.
The reconstructed mesh can be extracted from the fused TSDF volume with the Marching Cubes algorithm \cite{lorensenMarchingCubesHigh1987}.
This depth-based pipeline has two major drawbacks.
First, since single-view depth maps are estimated individually on each key frame, 
each depth estimation is from scratch instead of conditioned on the previous estimations even the view-overlapping is substantial.
As a result, the scale-factor may vary even with the correct camera ego-motion.
Due to depth inconsistencies between different views, the reconstruction result is prone to be either layered or scattered.
One example is shown in the red boxes in Fig. \ref{fig:teaser}, where the depth-based method struggles to produce coherent depth estimations on the chairs and wall.
Second, since key-frame depth maps need to be estimated separately in overlapped local windows, 
geometry of the same 3D surface is estimated multiple times in different key frames, causing redundant computation.

In this paper, we propose a novel framework for real-time monocular reconstruction named \shortname that jointly reconstructs and fuses the 3D geometry directly in the volumetric TSDF representation.
Given a sequence of monocular images and their corresponding camera poses estimated by a SLAM system, 
\shortname incrementally reconstructs local geometry in a view-independent 3D volume instead of view-dependent depth maps.
Specifically, it unprojects the image features to form a 3D feature volume and then uses sparse convolutions to process the feature volume to output a sparse TSDF volume.
With a coarse-to-fine design, the predicted TSDF is gradually refined at each level.
By directly reconstructing the implicit surface (TSDF), the network is able to 
learn the local smoothness and global shape  prior  of natural 3D surfaces. 
Different from depth-based methods that predict depth maps for each key frame separately, the surface geometry within a local fragment window is jointly predicted in \shortname, 
and thus \textit{locally} coherent geometry estimation can be produced.
To make the current-fragment reconstruction to be \textit{globally} consistent with the previously reconstructed fragments, 
a learning-based TSDF fusion module using the Gated Recurrent Unit (GRU) is proposed.
The GRU fusion makes the current-fragment reconstruction conditioned on the previously reconstructed global volume, 
yielding a joint reconstruction and fusion approach.
As a result, the reconstructed mesh is dense, accurate and globally coherent in scale.
Furthermore, predicting the volumetric representation also removes the redundant computation in depth-based methods, 
which allows us to use a larger 3D CNN while maintaining the real-time performance.

We validate our system on the ScanNet and 7-Scenes datasets. 
The experimental results show that
\shortname outperforms multiple state-of-the-art multi-view depth estimation methods and the volume-based reconstruction method Atlas~\cite{murezAtlasEndtoEnd3D2020} by a large margin, %
while achieving a real-time performance at 33 key frames per second, ${\sim}10\times$ faster compared to Atlas.
As shown in the supplementary video, our method is able to reconstruct large-scale 3D scenes from a video stream on a laptop GPU in real-time.
To the best of our knowledge, this is the first learning-based system that is able to reconstruct dense and coherent 3D scene geometry in real-time.

\section{Related Work}\label{sec:related}
\PAR{Multi-view Depth Estimation.}
The most related line of research is \textit{real-time methods} for multi-view depth estimation.
Before the age of deep learning, many renowned works in monocular 3D reconstruction \cite{vogiatzisVideobasedRealtimeMultiview2011,kolevTurningMobilePhones2014,schops3DModelingGo2015,pizzoliREMODEProbabilisticMonocular2014} have achieved good performance with plane-sweeping stereo and depth filters under the assumption of photo-consistency.
\cite{valentinDepthMotionSmartphone2019,yang2020Mobile3DReconRM} optimize this line of research towards low power consumption on mobile platforms.
Learning-based methods on real-time multi-view depth estimation try to alleviate the photo-consistency assumption with a data-driven approach.
Notably, MVDepthNet \cite{wangMVDepthNetRealtimeMultiview2018} and Neural RGB-$>$D \cite{liuNeuralRGBSensing2019} use 2D CNNs to process the 2D depth cost volume constructed from multi-view image features.
CNMNet \cite{longOcclusionAwareDepthEstimation2020} further leverages the planar structure in indoor scenes to constrain the surface normals calculated from the predicted depth maps to obtain smooth depth estimation.
These learning-based methods use 2D CNNs to process the depth cost volume to maintain a low computational cost for near real-time performance. %

When the input images are high-resolution and offline computation is allowed, multi-view depth estimation is also known as the \textit{Multiple View Stereo (MVS)} problem.
PatchMatch-based methods \cite{zhengPatchMatchBasedJoint2014,schonbergerPixelwiseViewSelection2016} have achieved impressive accuracy 
and are still the most popular methods applicable to high-resolution images.
Learning-based approaches in MVS have recently dominated several benchmarks \cite{aanaesLargeScaleDataMultipleView2016,knapitschTanksTemplesBenchmarking2017} in terms of accuracy, 
but are only limited to processing mid-resolution images due to the GPU memory constraint.
Different from the real-time methods, 3D cost volumes are constructed and 3D CNNs are used to process the cost volume as proposed in MVSNet \cite{yaoMVSNetDepthInference2018}.
Some recent works \cite{guCascadeCostVolume2019,chengDeepStereoUsing2019} improve this pipeline with a coarse-to-fine approach.
Similar design can also be found in many learning-based SLAM systems
 \cite{ummenhoferDeMoNDepthMotion2017,zhouDeepTAMDeepTracking2018,tangBANetDenseBundle2018,teedDeepV2DVideoDepth2018}.

All the above-mentioned works adopt single-view depth maps as intermediate representations. 
SurfaceNet \cite{jiSurfaceNetEndtoend3D2017,jiSurfaceNetEndtoend3D2020} takes a different approach and uses a unified \textit{volumetric representation} to predict the volume occupancy.
Recently, Atlas \cite{murezAtlasEndtoEnd3D2020} also proposes a volumetric design and direct predicts TSDF and semantic labels with 3D CNN.
As an offline method, Atlas aggregates the image features of the entire sequence and then predicts the global TSDF volume only once with a decoder module.
We further elaborate the relationship between the proposed method and Atlas in the supplementary material.
The proposed method is also related to \cite{choy3DR2N2UnifiedApproach2016,karLearningMultiViewStereo} in terms of using recurrent networks for multi-view feature fusion.
However, their recurrent fusion is applied to only the global features and their focus is to reconstruct single objects.

\begin{figure*}[ht]
    \vspace{-1.2cm}
    \centering
       \includegraphics[width=\linewidth]{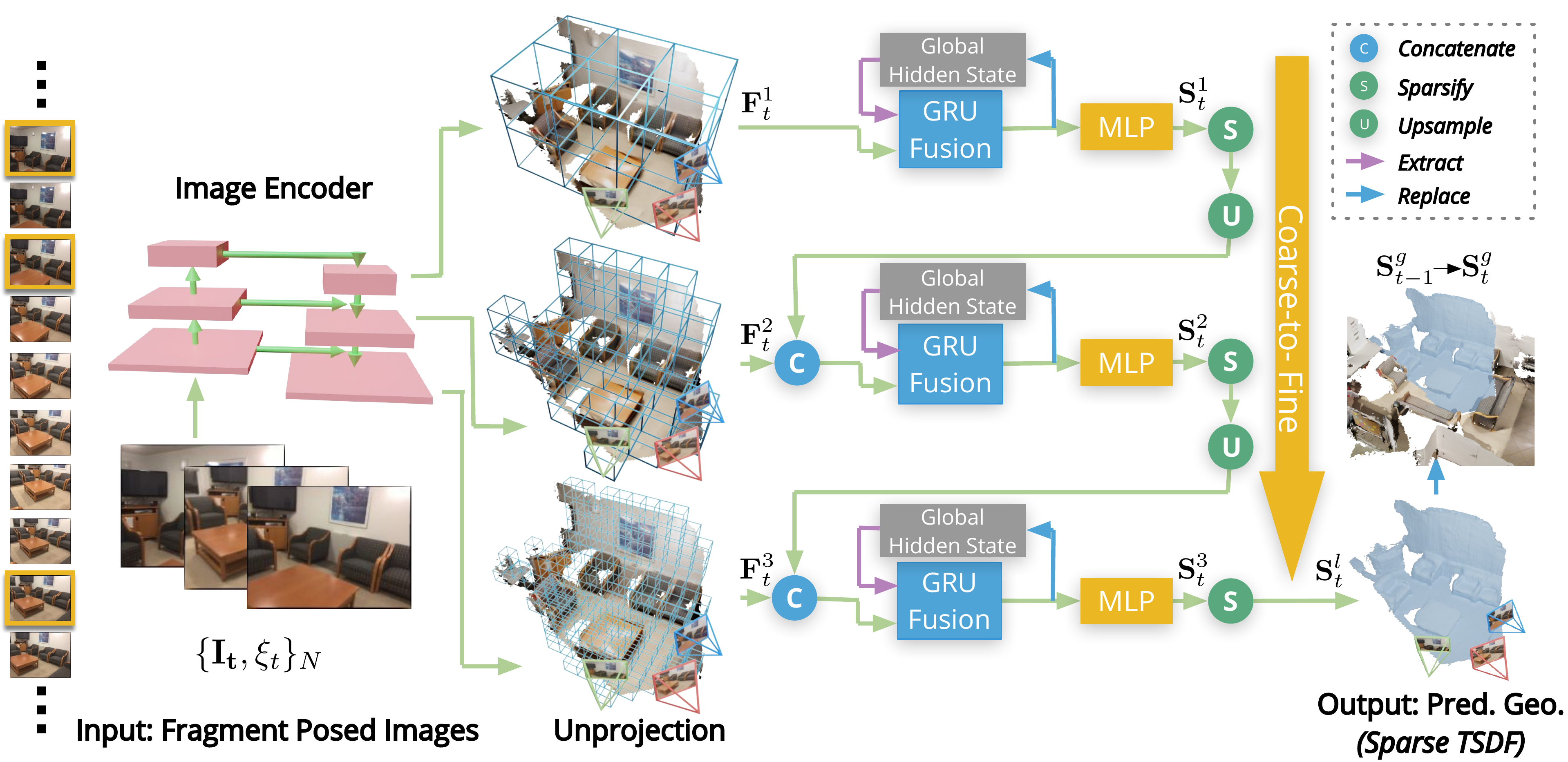}
       \caption{
           \textbf{\shortname architecture.}
           \shortname predicts TSDF with a three-level coarse-to-fine approach that gradually increases the density of sparse voxels.
           Key-frame images in the local fragment are first passed through the image backbone to extract the multi-level features. 
           These image features are later back-projected along each ray and aggregated into a 3D feature volume $\mathbf{F}_t^l$, where $l$ represents the level index. 
           At the first level ($l=1$), a dense TSDF volume $\mathbf{S}_t^{1}$ is predicted.
           At the second and third levels, the upsampled $\mathbf{S}_t^{l-1}$ from the last level is concatenated with $\mathbf{F}_t^l$ 
           and used as the input for the \textbf{\color{ProcessBlue}GRU Fusion} and \textbf{\color{YellowOrange}MLP} modules.
           A feature volume defined in the world frame is maintained at each level as the global hidden state of the GRU.
           At the last level, the output $\mathbf{S}_t^l$ is used to replace corresponding voxels in the global TSDF volume $\mathbf{S}_t^{g}$, 
           yielding the final reconstruction at time $t$. 
           }
       \label{fig:arch}
\end{figure*}
\PAR{3D Surface Reconstruction.}
After depth maps are estimated and converted to point clouds, the remaining task for 3D reconstruction is to estimate the 3D surface position and produce the reconstructed mesh.
In an offline MVS pipeline \cite{schonbergerPixelwiseViewSelection2016}, Poisson reconstruction \cite{kazhdanScreenedPoissonSurface2013} and Delaunay triagulation \cite{labatutRobustEfficientSurface2009} are often used to fulfill this purpose.
Proposed by the seminal work KinectFusion \cite{newcombeKinectFusionRealTimeDense2011}, 
incremental volumetric TSDF fusion \cite{curlessVolumetricMethodBuilding1996} gets widely adopted in real-time reconstruction scenarios due to its simplicity and parallelization capability.
\cite{niessnerRealtime3DReconstruction2013,daiBundleFusionRealtimeGlobally2016} improve KinectFusion by making it more scalable and robust.
RoutedFusion~\cite{wederRoutedFusionLearningRealtime2020,weder2020neuralfusion} changes the fusion operation from a simple linear addition into a data-dependent process.

\PAR{Neural Implicit Representations.}
Recently, neural implicit representations~\cite{meschederOccupancyNetworksLearning2018,parkDeepSDFLearningContinuous2019,saitoPIFuPixelAlignedImplicit2019,jiangLocalImplicitGrid2020,yarivMultiviewNeuralSurface2020,liu2020neural} have gained significant advances. 
Our work also learns a neural implicit representation by predicting SDF with the neural network from the encoded image features similar to PIFu~\cite{saitoPIFuPixelAlignedImplicit2019}.
The key difference is that we are using sparse 3D convolution to predict a discrete TSDF volume, instead of querying the MLP network with image features and 3D coordinates.

\section{Methods}\label{sec:method}
Given a sequence of monocular images $\{\mathbf{I}_t\}$ and camera pose trajectory $\{\xi_{t}\} \in \mathbb{SE}(3)$ provided by a SLAM system,
the goal is to reconstruct dense 3D scene geometry accurately in real-time.
We denote the global TSDF volume to reconstruct as $\mathbf{S}_t^g$, where $t$ represents the current time step. 
The system architecture is illustrated in Fig. \ref{fig:arch}.

\subsection{Key Frame Selection}\label{sec:keyframe}
To achieve real-time 3D reconstruction that is suitable for interactive applications, 
the reconstruction process needs to be incremental and the input images should be processed sequentially in local fragments \cite{choiRobustReconstructionIndoor}.
We seek to find a set of suitable key frames from the incoming image stream as input for the networks.
To provide enough motion parallax while keeping multi-view co-visibility for reconstruction, the selected key frames should be neither too close nor far from each other.
Following \cite{houMultiViewStereoTemporal2019}, a new incoming frame is selected as a key frame if its relative translation is greater than $t_{max}$ and the relative rotation angle is greater $R_{max}$.
A window with $N$ key frames is defined as a local fragment.
After key frames are selected, a cubic-shaped fragment bounding volume (FBV) that encloses all the key frame view-frustums is computed with a fixed max depth range $d_{max}$ in each view.
Only the region within the FBV is considered during the reconstruction of each fragment.

\begin{figure*}[ht]
    \vspace{-1.2cm}
    \centering
       \includegraphics[width=\linewidth]{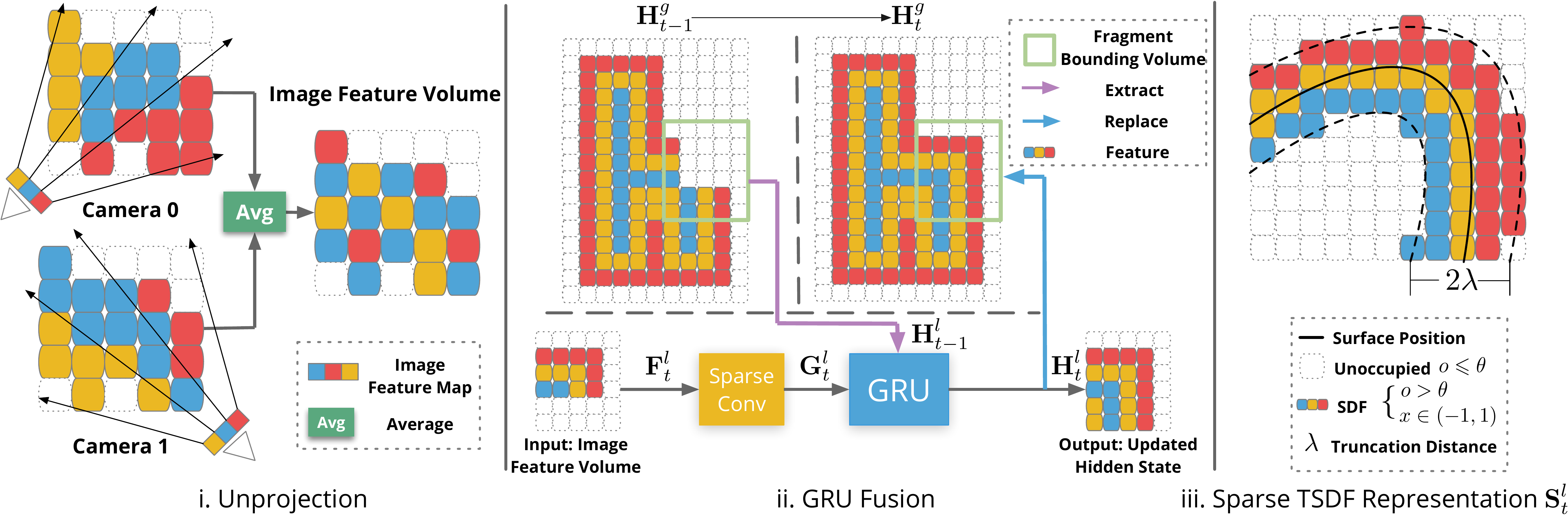}
       \caption{
           \textbf{2D toy examples to illustrate the unprojection, GRU fusion and sparse TSDF representation.}
           In figure i and ii, the colored grids mean different features.
           In figure iii, the colored grids mean different TSDF values.
           Best viewed in color.
           }
       \label{fig:toy_examples}
\end{figure*}
\subsection{Joint Fragment Reconstruction and Fusion}
We propose to simultaneously reconstruct the TSDF volume of a local fragment $\mathbf{S}_t^l$ and fuse it with global TSDF volume $\mathbf{S}_t^g$ with a learning-based approach.
The joint reconstruction and fusion is carried out in the local coordinate system.
The definition of the local and global coordinate systems as well as the construction of FBV are illustrated in Fig. 1 of the supplementary material.

\PAR{Image Feature Volume Construction.}
The $N$ images in the local fragment are first passed through the image backbone to extract the multi-level features.
Similar to previous works on volumetric reconstruction \cite{karLearningMultiViewStereo,jiSurfaceNetEndtoend3D2017,murezAtlasEndtoEnd3D2020}, 
the extracted features are back-projected along each ray into the 3D feature volume.
The image feature volume $\mathbf{F}_t^l$ is obtained by averaging the features from different views according to the visibility weight of each voxel.
The visibility weight is defined as the number of views from which a voxel can be observed in the local fragment.
A visualization of this unprojection process can be found in Fig.\ref{fig:toy_examples} i.

\PAR{Coarse-to-fine TSDF Reconstruction.}
We adopt a coarse-to-fine approach to gradually refine the predicted TSDF volume at each level.
We use 3D sparse convolution to efficiently process the feature volume $\mathbf{F}_t^l$.
The sparse volumetric representation also naturally integrates with the coarse-to-fine design.
Specifically, each voxel in the TSDF volume $\mathbf{S}_t^l$ contains two values, the occupancy score $o$ and the SDF value $x$.
At each level, both $o$ and $x$ are predicted by the MLP.
The occupancy score represents the confidence of a voxel being within the TSDF truncation distance $\lambda$.
The voxel whose occupancy score is lower than the sparsification threshold $\theta$ is defined as void space and will be sparsified.
This representation of sparse TSDF volume is visually illustrated in Fig.\ref{fig:toy_examples} iii.
After the sparsification, 
$\mathbf{S}_t^{l}$ is upsampled by $2\times$ and concatenated with the $\mathbf{F}_t^{l+1}$ as the input for the GRU Fusion module (introduced later) in the next level.

Instead of estimating single-view depth maps for each key frame, \shortname jointly reconstructs the implicit surface within the bounding volume of the local fragment window.
This design guides the network to learn the natural surface prior directly from the training data.
As a result, the reconstructed surface is locally smooth and coherent in scale.
Notably, this design also leads to less redundant computation compared to depth-based methods since each area on the 3D surface is estimated only once during the fragment reconstruction.

\PAR{GRU Fusion.}
To make the reconstruction consistent between fragments, 
we propose to make the current-fragment reconstruction to be conditioned on the reconstructions in previous fragments.
We use a 3D convolutional variant of Gated Recurrent Unit (GRU)~\cite{chungEmpiricalEvaluationGated2014} module for this purpose.
As illustrated in Fig.\ref{fig:toy_examples} ii, 
at each level the image feature volume $\mathbf{F}_t^l$ is first passed through the 3D sparse convolution layers to extract 3D geometric features $\mathbf{G}_t^l$.
The hidden state $\mathbf{H}_{t-1}^l$ is extracted from the global hidden state $\mathbf{H}_{t-1}^g$ within the fragment bounding volume. 
GRU fuses $\mathbf{G}_t^l$ with hidden state $\mathbf{H}_{t-1}^l$ and produces the updated hidden state $\mathbf{H}_t^l$, 
which will be passed through the MLP layers to predict the TSDF volume $\mathbf{S}_t^l$ at this level.
The hidden state $\mathbf{H}_{t}^l$ will also be updated to global hidden state $\mathbf{H}_{t}^g$ by directly replacing the corresponding voxels.
Formally, denoting $z_t$ as the update gate, $r_t$ as the reset gate, $\sigma$ as the sigmoid function and $W_{*}$ as the weight for sparse convolution,
GRU fuses $\mathbf{G}_t^l$ with hidden state $\mathbf{H}_{t-1}^l$ with the following operations:
\begin{align*}
    &z_t = \sigma(\text{SparseConv}([\mathbf{H}^l_{t-1}, \mathbf{G}_t^l], W_z)) \\
    &r_t = \sigma(\text{SparseConv}([\mathbf{H}^l_{t-1}, \mathbf{G}_t^l], W_r)) \\
    &\tilde{\mathbf{H}^l_{t}} = \tanh(\text{SparseConv}([r_t \odot \mathbf{H}^l_{t-1}, \mathbf{G}_t^l], W_h)) \\
    & \mathbf{H}^l_{t} = (1 - z_t) \odot \mathbf{H}^l_{t-1} + z_t \odot \tilde{\mathbf{H}^l_{t}}
\end{align*}

Intuitively, in the context of joint reconstruction and fusion of TSDF,
the update gate $z_t$ and forget gate $r_t$ in the GRU determine how much information from the previous reconstructions (i.e. hidden state $\mathbf{H}^l_{t-1}$) is fused to the current-fragment geometric feature $\mathbf{G}_t^l$, 
as well as how much information from the current-fragment will be fused into the hidden state $\mathbf{H}^l_{t}$.
As a data-driven approach, the GRU serves as a selective attention mechanism that replaces the linear running-average operation in conventional TSDF fusion~\cite{newcombeKinectFusionRealTimeDense2011}.
By predicting $\mathbf{S}_t^l$ after the GRU, 
the MLP network can leverage the context information accumulated from history fragments to produce consistent surface geometry across local fragments.
This is also conceptually analogous to the depth filter in a non-learning-based 3D reconstruction pipeline~\cite{schops3DModelingGo2015,pizzoliREMODEProbabilisticMonocular2014}, 
where the current observation and the temporally-fused depths are fused with the Bayesian filter.
The effectiveness of joint reconstruction and fusion is validated in the ablation study.

\PAR{Integration to the Global TSDF Volume.}
At the last coarse-to-fine level, $\mathbf{S}_t^3$ is predicted and further sparsified to $\mathbf{S}_t^l$.
Since the fusion between $\mathbf{S}_t^l$ and $\mathbf{S}_t^g$ has been done in GRU Fusion, $\mathbf{S}_t^l$ is integrated into $\mathbf{S}_t^g$ by directly replacing the corresponding voxels after being transformed into the global coordinate.
At each time step $t$, Marching Cubes is performed on $\mathbf{S}_t^g$ to reconstruct the mesh.

\PAR{Supervision.}
Following \cite{daiSGNNSparseGenerative2019}, two loss functions are used to supervise the network. 
The occupancy loss is defined as the binary cross-entropy (BCE) between the predicted occupancy values and the ground-truth occupancy values.
The SDF loss is defined as the $\ell_1$ distance between the predicted SDF values and the ground-truth SDF values. 
We log-transform the SDF values of predictions and ground-truth before applying the $\ell_1$ loss.
The supervision is applied to all the coarse-to-fine levels.

\subsection{Implementation Details}
We use \texttt{torchsparse} \cite{tangSearchingEfficient3D2020} as the implementation of 3D sparse convolution.
The image backbone is a variant of MnasNet \cite{tan2019mnasnet} and is initialized with the weights pretrained from ImageNet.
Feature Pyramid Network \cite{linFeaturePyramidNetworks2017} is used in the backbone to extract more representative multi-level features. 
The entire network is trained end-to-end with randomly initialized weights except for the image backbone.
The occupancy score $o$ is predicted with a Sigmoid layer.
The voxel size of the last level is $4cm$ and the TSDF truncation distance $\lambda$ is set to $12cm$.
$d_{max}$ is set to $3m$.
$R_{max}$ and $t_{max}$ are set to 15\degree and $0.1m$ respectively.
$\theta$ is set to $0.5$.
Nearest-neighbor interpolation is used in the upsampling between coarse-to-fine levels.

\section{Experiments}\label{sec:exp}
In this section, we conduct a series of experiments to evaluate the reconstruction quality and different design considerations of \shortname.

\subsection{Datasets, Metrics, Baselines and Protocols.}\label{sec:dmb}
\PAR{Datasets.} We perform the experiments on two indoor datasets, ScanNet (V2)~\cite{daiScanNetRichlyannotated3D2017} and 7-Scenes~\cite{shottonSceneCoordinateRegression2013}.
The ScanNet dataset contains 1613 indoor scenes with ground-truth camera poses, surface reconstructions, and semantic segmentation labels.
There are two training/validation splits commonly used in previous works (defined in \cite{murezAtlasEndtoEnd3D2020} and \cite{tangBANetDenseBundle2018}) for the ScanNet dataset.
We use the same training and validation data with the corresponding baseline methods to make a fair comparison.
The 7-Scenes dataset is another challenging RGB-D dataset captured in indoor scenes.
Following the baseline method~\cite{longOcclusionAwareDepthEstimation2020}, we use the model trained on ScanNet to perform the validation on 7-Scenes.

\PAR{Metrics.} The 3D reconstruction quality is evaluated using 3D geometry metrics presented in \cite{murezAtlasEndtoEnd3D2020}, as well as standard 2D depth metrics defined in \cite{eigenDepthMapPrediction2014}.
The definitions of these metrics are detailed in the supplementary material.
Among these 3D and 2D metrics, we consider \textit{F-score} as the most suitable metrics to measure 3D reconstruction quality since both the accuracy and completeness of the reconstruction are considered. 

\PAR{Baselines.} We compare our method with the following baseline methods in three categories: 
1) \textit{Real-time methods} for multi-view depth estimation \cite{wangMVDepthNetRealtimeMultiview2018,houMultiViewStereoTemporal2019,liuNeuralRGBSensing2019,longOcclusionAwareDepthEstimation2020}. 
Due to the efficiency constraints, the estimated depth accuracy by these methods is rather limited.
We compare with these methods to demonstrate the better reconstruction accuracy of \shortname given the same efficiency.
2) \textit{Multiple View Stereo} methods \cite{schonbergerPixelwiseViewSelection2016,imDPSNetEndtoendDeep2019,yaoMVSNetDepthInference2018,murezAtlasEndtoEnd3D2020,luoConsistentVideoDepth2020}. 
These offline methods have much higher accuracy compared to real-time methods.
These baselines are used to demonstrate that \shortname achieves a reconstruction quality on-par with offline methods but runs in real-time.
3) \textit{Learning-based SLAM} methods \cite{ummenhoferDeMoNDepthMotion2017,tangBANetDenseBundle2018,teedDeepV2DVideoDepth2018}.
These monocular SLAM methods estimate camera poses and perform reconstruction simultaneously, thus the scale factor of pose and depth is usually not accurately estimated.
For a fair comparison, we use ground-truth camera poses for these methods and apply a scaling factor to the predicted depth map using ground-truth depth.
Among all these baseline methods, \textit{GPMVS}~\cite{houMultiViewStereoTemporal2019} and \textit{Atlas}~\cite{murezAtlasEndtoEnd3D2020} are the most relevant real-time and offline methods, respectively.

\PAR{Evaluation Protocols.} Since our method does not estimate depth maps explicitly, we render the reconstructed mesh to the image plane and obtain depth map estimations \cite{murezAtlasEndtoEnd3D2020}.
Key frames used for evaluation are sampled from the video sequence with an interval of 10 frames for both depth-based methods and Atlas.
Following \cite{murezAtlasEndtoEnd3D2020,longOcclusionAwareDepthEstimation2020}, \cite{yaoMVSNetDepthInference2018,wangMVDepthNetRealtimeMultiview2018,imDPSNetEndtoendDeep2019,houMultiViewStereoTemporal2019} are fine-tuned on ScanNet.
To evaluate depth-based methods \cite{schonbergerPixelwiseViewSelection2016, wangMVDepthNetRealtimeMultiview2018, houMultiViewStereoTemporal2019, imDPSNetEndtoendDeep2019} in 3D, 
we use the point cloud fusion to obtain the 3D reconstruction following Atlas.
For other depth-based methods, we use the standard TSDF fusion proposed in~\cite{newcombeKinectFusionRealTimeDense2011,curlessVolumetricMethodBuilding1996}. 
For the reasons we detailed in the supplementary material, in order to make a fair comparison with Atlas, we also report the evaluation results using the double-layered mesh (same as Atlas).
The evaluation of 3D geometry on 7-Scenes uses the single-layered mesh.
We also evaluate the depth filtering operation with multi-view consistency check, which will be elaborated in the supplementary material.
\begin{table*}[ht]
    \vspace{-1.2cm}
    \centering
    \resizebox{0.75\textwidth}{!}{
    \begin{tabular}{cccccc>{\columncolor[gray]{0.902}}cc}
    \Xhline{3\arrayrulewidth}
    Method                                                 & Layer               & Comp $\downarrow$ & Acc $\downarrow$           & Recall $\uparrow$         & Prec $\uparrow$           & \textbf{F-score} $\uparrow$        & Time ($ms$) $\downarrow$   \\ \hline
    MVDepthNet \cite{wangMVDepthNetRealtimeMultiview2018}  & single                                          & 0.040           & 0.240           & 0.831           & 0.208           & 0.329           & 48     \\ 
    GPMVS \cite{houMultiViewStereoTemporal2019}            & single                                     & \textbf{0.031}  & 0.879           & \textbf{0.871}  & 0.188           & 0.304           & 51     \\ 
    DPSNet \cite{imDPSNetEndtoendDeep2019}                 & single                               & 0.045           & 0.284           & 0.793           & 0.223           & 0.344           & 322    \\ 
    COLMAP \cite{schonbergerPixelwiseViewSelection2016}    & single      & 0.069           & 0.135           & 0.634           & 0.505           & 0.558           & 2076   \\ 
    Ours                                                   & single      & 0.128           & \textbf{0.054}  & 0.479           & \textbf{0.684}  & \textbf{0.562}           & \textbf{30}  \\
    \hline
    Atlas \cite{murezAtlasEndtoEnd3D2020}                  & double     & \textbf{0.062}           & 0.128           & \textbf{0.732}         & 0.382           & 0.499           & 292   \\ 
    Ours                                                   & double      & 0.106           & \textbf{0.073}  & 0.609           & \textbf{0.450}  & \textbf{0.516}  & \textbf{30}     \\ 
    \Xhline{3\arrayrulewidth}
    DeepV2D \cite{teedDeepV2DVideoDepth2018}               & single                                & \textbf{0.057} & 0.239          & \textbf{0.646} & 0.329          & 0.431          & 347    \\ 
    Consistent Depth \cite{luoConsistentVideoDepth2020}    & single                                  & 0.091          & 0.344          & 0.461          & 0.266          & 0.331          & 2321   \\ 
    Ours                                                   & single      & 0.120          & \textbf{0.062} & 0.428          & \textbf{0.592} & \textbf{0.494} & \textbf{30}    \\ 
    \Xhline{3\arrayrulewidth}
    \end{tabular}
    }
    \caption{\textbf{3D geometry metrics on ScanNet.} 
    We use two different training/validation splits following Atlas \cite{murezAtlasEndtoEnd3D2020} (top block) and  BA-Net \cite{tangBANetDenseBundle2018} (bottom block).
    We elaborate the meaning of the single and double layer in the supplementary material.}
    \label{tab:scannet-3d}
\end{table*}
\begin{table*}[ht]
    \centering
    \resizebox{0.75\textwidth}{!}{
    \begin{tabular}{ccccccccc}
    \Xhline{3\arrayrulewidth}
    Method                                        & Abs Rel $\downarrow$ & Abs Diff $\downarrow$ & Sq Rel $\downarrow$ & RMSE $\downarrow$ & $\delta < 1.25$ $\uparrow$& Comp $\uparrow$ \\ \hline
    COLMAP \cite{schonbergerPixelwiseViewSelection2016}                                        & 0.137          & 0.264          & 0.138          & 0.502          & 83.4                    & 0.871           \\ 
    MVDepthNet \cite{wangMVDepthNetRealtimeMultiview2018}                                   & 0.098          & 0.191          & 0.061          & 0.293          & 89.6                    & 0.928           \\ 
    GPMVS \cite{houMultiViewStereoTemporal2019}                                        & 0.130          & 0.239          & 0.339          & 0.472          & 90.6                    & 0.928           \\ 
    DPSNet \cite{imDPSNetEndtoendDeep2019}                                       & 0.087          & 0.158          & 0.035          & 0.232          & 92.5           & 0.928           \\ 
    Atlas \cite{murezAtlasEndtoEnd3D2020}                                        & 0.065          & 0.123          & 0.045          & 0.251          & 93.6                    & \textbf{0.999}  \\ 
    Ours                                          & \textbf{0.065} & \textbf{0.106} & \textbf{0.031} & \textbf{0.195} & \textbf{94.8}           & 0.909           \\ 
    \Xhline{3\arrayrulewidth}

    Method                                & Abs Rel $\downarrow$ & Sq Rel $\downarrow$ & RMSE $\downarrow$ & RMSE log $\downarrow$ & Sc Inv $\downarrow$ &- \\ \hline
    DeMoN \cite{ummenhoferDeMoNDepthMotion2017}                                & 0.231          & 0.520          & 0.761          & 0.289          & 0.284          &- \\ 
    BA-Net \cite{tangBANetDenseBundle2018}                               & 0.161          & 0.092          & 0.346          & 0.214          & 0.184          &- \\ 
    DeepV2D \cite{teedDeepV2DVideoDepth2018}                              & 0.057          & \textbf{0.010} & 0.168          & \textbf{0.080} & \textbf{0.077} &- \\ 
    Consistent Depth \cite{luoConsistentVideoDepth2020}                     & 0.073          & 0.037          & 0.217          & 0.105          & 0.103          &- \\ 
    Ours                                  & \textbf{0.047} & 0.024          & \textbf{0.164} & 0.093          & 0.092          &- \\ 
    \Xhline{3\arrayrulewidth}
    \end{tabular}
    }
    \caption{\textbf{2D depth metrics on ScanNet.} 
    We use two different training/validation splits following Atlas \cite{murezAtlasEndtoEnd3D2020} (top block) and  BA-Net \cite{tangBANetDenseBundle2018} (bottom block).
    }
    \vspace{-0.2cm}
    \label{tab:scannet-2d}
 \end{table*}

\subsection{Evaluation Results}\label{sec:eval_results}

\PAR{ScanNet.}
2D depth metrics and 3D geometry metrics are used on the ScanNet dataset.
The 3D geometry evaluation results are shown in Tab. \ref{tab:scannet-3d}. 
Our method produces much better performance than recent learning-based methods and achieves slightly better results than COLMAP. 
We believe that the improvements come from the joint reconstruction and fusion design achieved by the GRU Fusion module.
Compared to depth-based methods, \shortname can produce coherent reconstructions both locally and globally.
Our method also surpasses the volumetric baseline method Atlas \cite{murezAtlasEndtoEnd3D2020} on the accuracy, precision, and F-score.
The improvements potentially come from the design of local fragment separation in our method, 
which can act as a view-selection mechanism that avoids irrelevant image features to be fused into the 3D volume.
In terms of completeness and recall, the proposed method has an inferior performance compared to both depth-based methods and Atlas. 
Since depth-based methods predict pixel-wise depth maps on each view, the coverage of their predictions is high by nature, but with the cost of accuracy. 
Being an offline approach, Atlas has the advantage of having a global context from the entire sequence before predicting the geometry.
As a result, Atlas sometimes achieves even better completeness compared to the ground-truth due to its TSDF completion capability.
However, Atlas tends to predict over-smoothed geometries, and the completed regions may be inaccurate.
As for 2D depth metrics, \shortname also outperforms previous state-of-the-art methods for almost all 2D depth metrics, as shown in Tab. \ref{tab:scannet-2d}. %

\PAR{7-Scenes.}
2D depth metrics and 3D geometry metrics are evaluated on the 7-Scenes dataset.
As shown in Tab. \ref{tab:7scenes-2d}, our method achieves comparable performance to the state-of-the-art method CNMNet \cite{longOcclusionAwareDepthEstimation2020} and outperforms all other methods. 
We believe that the accuracy of the proposed method can be further improved by leveraging the planar structure information as in CNMNet.
Since the model used here is only trained on ScanNet, the results also demonstrate that \shortname can generalize well beyond the domain of the training data.

\PAR{Efficiency.}
We also report the average running time of the baselines and our method in Tab. \ref{tab:scannet-3d}. 
Only the inference time on key frames is computed.
A detailed timing analysis for each module of \shortname is presented in Table \ref{tab:time}.
For volumetric methods (Atlas and ours), the running time is obtained by dividing the time of reconstructing the TSDF volume of a local fragment by the number of key frames in the local fragment.
Notice that the time for TSDF fusion is not included for depth-based methods. 
The running time for \cite{teedDeepV2DVideoDepth2018,luoConsistentVideoDepth2020,liuNeuralRGBSensing2019,longOcclusionAwareDepthEstimation2020, ummenhoferDeMoNDepthMotion2017} and \shortname is measured on an NVIDIA RTX 2080Ti GPU. 
We use running time reported in \cite{murezAtlasEndtoEnd3D2020} and \cite{yu2020fast} for \cite{wangMVDepthNetRealtimeMultiview2018,imDPSNetEndtoendDeep2019,schonbergerPixelwiseViewSelection2016,houMultiViewStereoTemporal2019,murezAtlasEndtoEnd3D2020} and \cite{yaoMVSNetDepthInference2018}, respectively.
\begin{table}[ht]
    \centering
    \resizebox{1.0\textwidth}{!}{
        \begin{tabular}{ccccccc}
            \Xhline{3\arrayrulewidth}
            Method                                              & Comp $\downarrow$                      & Acc $\downarrow$     & Recall $\uparrow$   & Prec $\uparrow$   & \textbf{F-score} $\uparrow$ \\
            \hline
            DeepV2D \cite{teedDeepV2DVideoDepth2018}            & 0.180                                  & 0.518                & 0.175               & 0.087             & 0.115                       \\
            CNMNet \cite{longOcclusionAwareDepthEstimation2020} & \textbf{0.150}                         & 0.398                & \textbf{0.246}      & 0.111             & 0.149                       \\
            Ours                                                & 0.228                                  & \textbf{0.100}       & 0.227               & \textbf{0.389}    & \textbf{0.282}              \\
            \Xhline{3\arrayrulewidth}
            Method                                              & \footnotesize$\delta < 1.25 \uparrow$ & Abs Rel $\downarrow$ & Sq Rel $\downarrow$ & RMSE $\downarrow$ & Time  $\downarrow$           \\ \hline
            DeMoN \cite{ummenhoferDeMoNDepthMotion2017}         & 31.88                                  & 0.3888               & 0.4198              & 0.8549            & 110                         \\
            MVSNet \cite{yaoMVSNetDepthInference2018}           & 64.09                                  & 0.2339               & 0.1904              & 0.5078            & 1050                        \\
            N-RGBD \cite{liuNeuralRGBSensing2019}               & 69.26                                  & 0.1758               & 0.1123              & 0.4408            & 202                         \\
            MVDNet \cite{wangMVDepthNetRealtimeMultiview2018}   & 71.79                                  & 0.1925               & 0.2350              & 0.4585            & 48                          \\
            DPSNet \cite{imDPSNetEndtoendDeep2019}              & 70.96                                  & 0.1991               & 0.1420              & 0.4382            & 322                         \\
            DeepV2D \cite{teedDeepV2DVideoDepth2018}            & 42.80                                  & 0.4370               & 0.5530              & 0.8690            & 347                         \\
            CNMNet \cite{longOcclusionAwareDepthEstimation2020} & 76.64                                  & 0.1612               & \textbf{0.0832}     & 0.3614            & 80                          \\
            Ours                                                & \textbf{82.00}                         & \textbf{0.1550}      & 0.1040              & \textbf{0.3470}   & \textbf{30}                 \\
            \Xhline{3\arrayrulewidth}
        \end{tabular} }
    \caption{\textbf{3D geometry metrics (top block) and 2D depth metrics (bottom block) on 7-Scenes.} Time is measured in milliseconds.}
    \label{tab:7scenes-2d}
\end{table}

As shown in Tab. \ref{tab:scannet-3d}, our time cost is 30ms per key frame,  achieving real-time speed at 33 key frames per second and outperforming all previous methods.
Specifically, our method runs $\sim$\textbf{10$\times$ faster} than Atlas, and \textbf{77$\times$ faster} than Consistent Depth.
Predicting the volumetric representation removes the redundant computation in depth-based methods, which contributes to the fast running speed of our method.
Compared to Atlas, incrementally reconstructing geometry in local fragment avoids processing a huge 3D volume, leading to a faster speed than Atlas.
The use of sparse convolution also contributes to the superior efficiency of \shortname.

\vspace{-0.1cm}
\subsection{Ablation Study}\label{sec:ablation}
In this section, we conduct several ablation experiments on the ScanNet dataset to discuss the effectiveness of components in our method.

\PAR{GRU Fusion.}
We validate the GRU Fusion design by comparing rows from (i) to (iv) in Tab. \ref{tab:ablation}.

To validate the benefit of feature fusion, we compare row (i) and row (ii) in Tab. \ref{tab:ablation}. Using feature fusion with the average operation obtains nearly 5\% improvement for the precision metric than conventional linear TSDF fusion.
Visualization in Fig. \ref{fig:ablation} shows that feature fusion with the average operation can reconstruct smoother geometry.
These results demonstrate that feature fusion can be more effective than TSDF fusion using the same average operation.

Comparing row (ii) and row (iii) in Tab. \ref{tab:ablation} shows that replacing average operation with GRU gives 4\% improvement in terms of recall. 
The mesh in Fig. \ref{fig:ablation} (iii) is also more complete than that in Fig. \ref{fig:ablation} (ii). 
These results demonstrate that the GRU is more effective to 
selectively integrate \textit{only} the consistent information from the current-fragment to the hidden state.

The recalls in row (iii) and row (iv) in Tab. \ref{tab:ablation} show that fusion in the fragment bounding volume can produce much more complete results.
Visualization results in Fig. \ref{fig:ablation} (iii) and (iv) show that, with fusion in the fragment bounding volume, our method produces fewer artifacts on the ground.
Fusion in the fragment bounding volume can leverage the context information in boundaries and produce more consistent and complete surface estimation.

\begin{table}[]
    \centering
    \resizebox{1.0\textwidth}{!}{
    \begin{tabular}{cccccc}
    \Xhline{3\arrayrulewidth}
    Img. Enc.                  &              & Unproj.       & Sparse Conv. & GRU    & Total                      \\ \hline
    \multirow{3}{*}{4.03}          & Level 1      & 1.27         & 3.70         & 2.18    & \multirow{3}{*}{29.56}    \\ \cline{2-5}
                                   & Level 2      & 1.21         & 3.84         & 2.24    &                           \\ \cline{2-5}
                                   & Level 3      & 2.18         & 5.11         & 3.80        &                      \\ \cline{2-5}
    \Xhline{3\arrayrulewidth}
    \end{tabular}
    }
    \caption{\textbf{Timing analysis} of \shortname measured in milliseconds per key~frame. The level number indicates the different coarse-to-fine level. Img. Enc. stands for image encoder, Unproj. stands for unprojection.}
    \label{tab:time}
\end{table}
\begin{table}[ht]
    \centering
    \resizebox{1.0\textwidth}{!}{
    \begin{tabular}{ccccccc}
    \Xhline{3\arrayrulewidth}
                  & \multirow{2}{*}{\#views}                & \multicolumn{2}{c}{Fusion}           & \multicolumn{3}{c}{3D Geometry Metrics}                  \\ \cline{3-7}
                  &               & Area         & Method       & Recall          & Prec           & \textbf{F-score}        \\ \hline
    i             & 5             & OCC          & Linear       & 0.576          & 0.386          & 0.462                   \\ 
    ii            & 5             & OCC          & Avg          & 0.535          & 0.432          & 0.478                   \\ 
    iii           & 5             & OCC          & GRU          & 0.572          & 0.426          & 0.488                   \\ 
    iv            & 5             & FBV          & GRU          & \textbf{0.613} & 0.421          & 0.494                   \\ 
    -             & 7             & FBV          & GRU          & 0.607          & 0.435          & 0.507                   \\ 
    v             & 9             & FBV          & GRU          & 0.609          & \textbf{0.450} & \textbf{0.516}          \\ 
    -             & 11            & FBV          & GRU          & 0.593          & 0.398          & 0.474                   \\ 
    \Xhline{3\arrayrulewidth}
    \end{tabular}
    }
    \caption{\textbf{Ablation study.} We report 3D geometry metrics on ScanNet. 
    \textit{OCC}: fuse 3D geometric features $\mathbf{G}_t^l$ within the occupied area where occupancy score $o > \theta$. 
    \textit{FBV}: fuse 3D geometric features $\mathbf{G}_t^l$ within the fragment bounding volume.
    \textit{Linear}: remove GRU-Fusion and use the conventional running-average-based linear TSDF fusion to update the global TSDF volume. 
    \textit{Avg}: fuse 3D geometric features $\mathbf{G}_t^l$ with the average operation.
    \textit{GRU}: fuse 3D geometric features $\mathbf{G}_t^l$ with GRU.
    We use row (v) in all other experiments. 
    More details about ablation experiments can be found in the supplementary material.
}
    \label{tab:ablation}
\end{table}
\PAR{Number of views.}
We set 5, 7, 9 and 11 views as the length of a fragment respectively.
As shown in row (v) in Tab. \ref{tab:ablation}, the F-score has over 2\% improvement when 9 views are used as a fragment.
As shown in visualization results in Fig. \ref{fig:ablation} (v), with more views in a fragment, the geometry can be reconstructed more accurately compared to Fig. \ref{fig:ablation} (iv).

\begin{figure*}[ht]
    \vspace{-1.2cm}
    \centering
       \includegraphics[width=\linewidth]{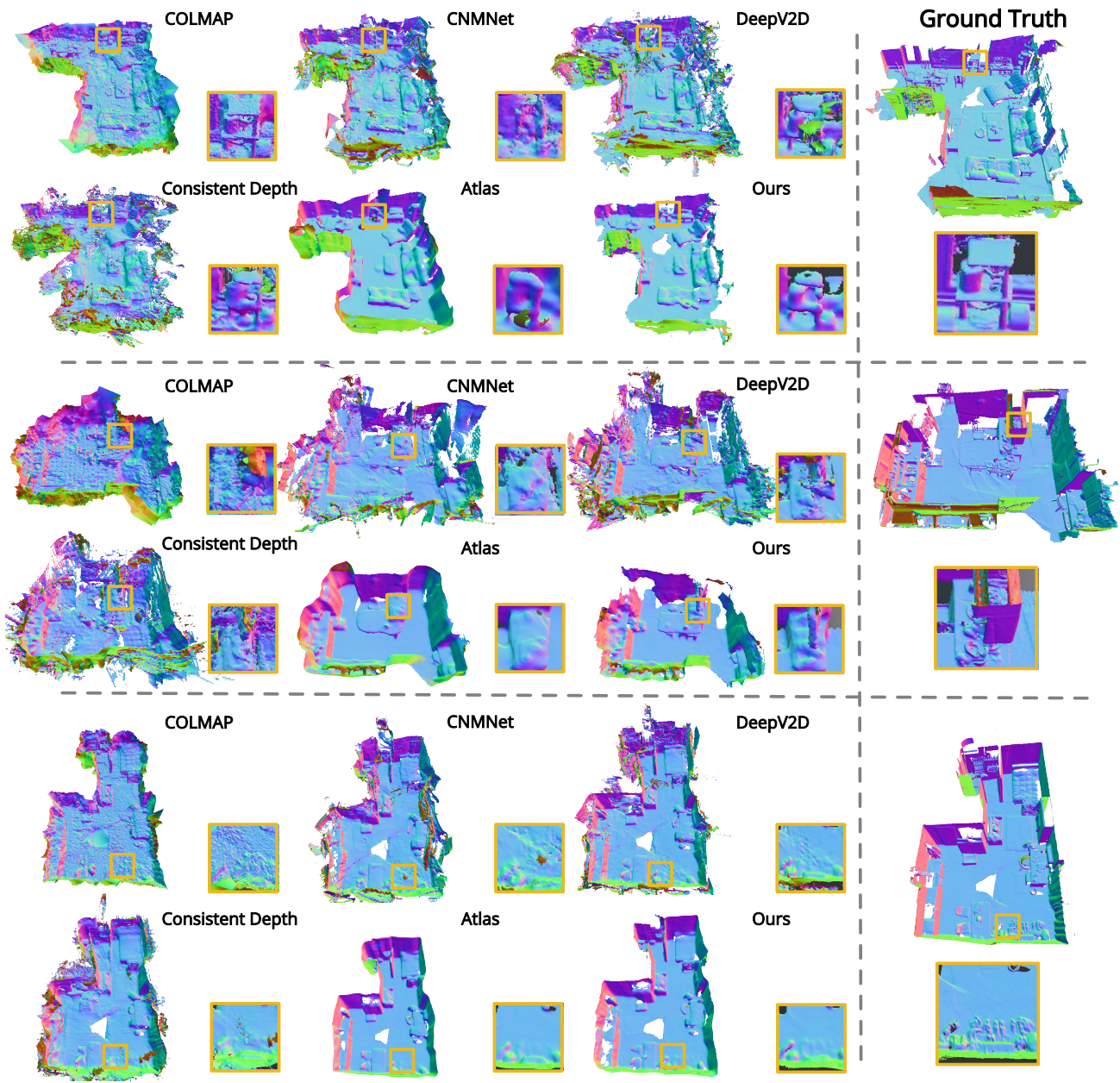}
       \caption{
           \textbf{Qualitative results on ScanNet.}
        Compared to depth-based methods, \shortname can produce much more coherent reconstruction results. 
        Notice that our method also recovers sharper geometry compared to Atlas \cite{murezAtlasEndtoEnd3D2020}, which illustrates the effectiveness of the local fragment design in our method. 
        Reconstructing only within the local fragment window avoids irrelevant image features from far-away camera views to be fused into the 3D volume. 
        The color indicates surface normal.
        More qualitative results can be found in the supplementary material and the \href[pdfnewwindow=true]{https://zju3dv.github.io/neuralrecon/}{project webpage}.
        Zoom in for details.}
       \label{fig:qualitative}
\end{figure*}

\begin{figure*}[h]
    \centering
       \includegraphics[width=\linewidth]{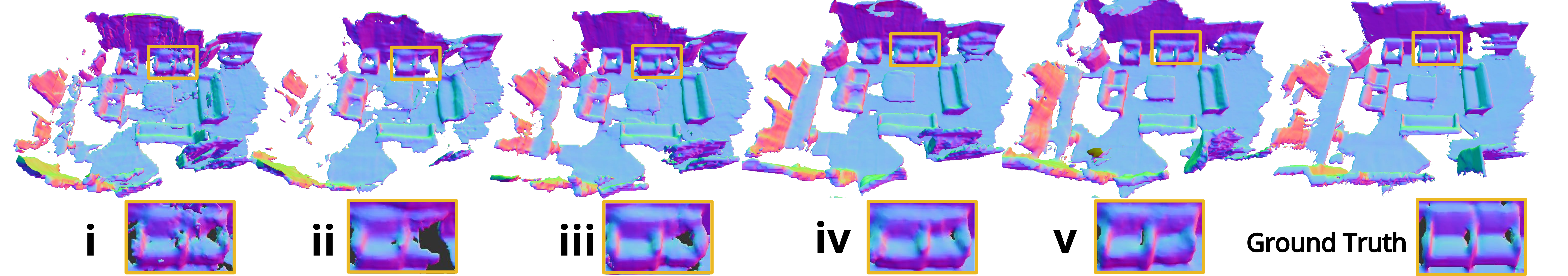}
       \caption{
           \textbf{Ablation study.} The indications of Roman numerals are in Tab. \ref{tab:ablation}. The analysis is presented in Sec.  \ref{sec:ablation}.
           }
       \label{fig:ablation}
\end{figure*}

\PAR{Qualitative Results.}\label{sec:qualitative_results}
We provide the qualitative results and the corresponding analysis in Fig. \ref{fig:qualitative}.

\vspace{-0.2cm}
\section{Conclusion}
In this paper, we introduced a novel system \shortname for real-time 3D reconstruction with monocular video.
The key idea is to jointly reconstruct and fuse sparse TSDF volumes for each video fragment incrementally by 3D sparse convolutions and GRU.
This design enables \shortname to output accurate and coherent reconstruction in real-time.
Experiments show that \shortname outperforms state-of-the-art methods in both reconstruction quality and running speed. 
The sparse TSDF volume reconstructed by \shortname can be directly used in downstream tasks like 3D object detection, 3D semantic segmentation and neural rendering.
We believe that, by jointly training with the downstream tasks end-to-end, \shortname enables new possibilities in learning-based multi-view perception and recognition systems.

\PAR{Acknowledgement.}
The authors would like to acknowledge the support from the National Key Research and Development Program of China (No. 2020AAA0108901), NSFC (No. 61806176), and ZJU-SenseTime Joint Lab of 3D Vision.

\clearpage

{\small
\bibliographystyle{ieee_fullname}
\bibliography{fulldb_cleaned}
}

\end{document}